\documentclass[10pt,twocolumn,letterpaper]{article}

\usepackage[pagenumbers]{cvpr} % To force page numbers, e.g. for an arXiv version

\usepackage{graphicx}
\usepackage{amsmath}
\usepackage{amssymb}
\usepackage{booktabs}
\usepackage{bbold}

\usepackage[pagebackref,breaklinks,colorlinks]{hyperref}
\usepackage{caption}

\usepackage[capitalize]{cleveref}
\crefname{section}{Sec.}{Secs.}
\Crefname{section}{Section}{Sections}
\Crefname{table}{Table}{Tables}
\crefname{table}{Tab.}{Tabs.}

\def\cvprPaperID{17} % *** Enter the CVPR Paper ID here
\def\confName{CVPR}
\def\confYear{2022}

\begin{document}

\title{Causality for Inherently Explainable Transformers: CAT-XPLAIN}
\author{
Subash Khanal$^{1,2}$~~~
Benjamin Brodie$^1$~~~
Xin Xing$^{1,2}$~~~
Ai-Ling Lin$^2$~~~
Nathan Jacobs$^1$~~~
\smallskip 
\\
$^1$Department of Computer Science, University of Kentucky, Lexington, KY, USA
\\
$^2$Department of Radiology, University of Missouri, Columbia, MO, USA\\
{\tt\small subash.khanal.cs@gmail.com}
}
\maketitle

%%%%%%%%% ABSTRACT
\begin{abstract}
   There have been several post-hoc explanation approaches developed to explain pre-trained black-box neural networks. However, there is still a gap in research efforts toward designing neural networks that are inherently explainable. In this paper, we utilize a recently proposed instance-wise post-hoc causal explanation method to make an existing transformer architecture inherently explainable. Once trained, our model provides an explanation in the form of top-$k$ regions in the input space of the given instance contributing to its decision. We evaluate our method on binary classification tasks using three image datasets: MNIST, FMNIST, and CIFAR. Our results demonstrate that compared to the causality-based post-hoc explainer model, our inherently explainable model achieves better explainability results while eliminating the need of training a separate explainer model. Our code is available at \url{https://github.com/mvrl/CAT-XPLAIN}.
\end{abstract}

%%%%%%%%% BODY TEXT
\section{Introduction}
\label{sec:intro}

Explainable AI (XAI) aims at designing methods or explainers that provide reasoning for the decisions made by a model trained for a specific task. XAI approaches can be broadly grouped under two categories: post-hoc explainers and explanation through inherently explainable models. Post-hoc approaches use backpropagation-based techniques~\cite{Grad-CAM}%
, perturbation-based methods ~\cite{Occlusion_Sensitivity},
or train a post-hoc explainer model ~\cite{schwab2019cxplain}
to highlight regions of the input instance considered important for the decision of a pre-trained black box model.

Post-hoc explanation methods are often model-agnostic and do not affect the black-box model's performance during the explanation. However, there can be differences in inductive bias between the black-box and the post-hoc explainer. Moreover, the post-hoc explainers are trained in isolation while being guided by the output of a pre-trained black box. These explainers can be learning different feature representations and focus on different input regions than the black box. Therefore, there have been raising concerns regarding the faithfulness of post-hoc explanations~\cite{rudin2019stop}. Accordingly, there has been a push towards developing inherently explainable models for high-stakes decisions such as AI-based medical diagnosis, biomarker discovery, etc. In this line of thinking, we propose a small modification to an existing vision transformer architecture~\cite{dosovitskiy2020image} and its training to build an inherently explainable model which identifies the most causally significant input regions that contribute to its decision.
%-------------------------------------------------------------------------

\section{Causality based post-hoc explanation}
\label{sec:Causality}
\begin{figure}[htbp]
    \centering
    \includegraphics[width=0.8\linewidth,height=0.5\linewidth]{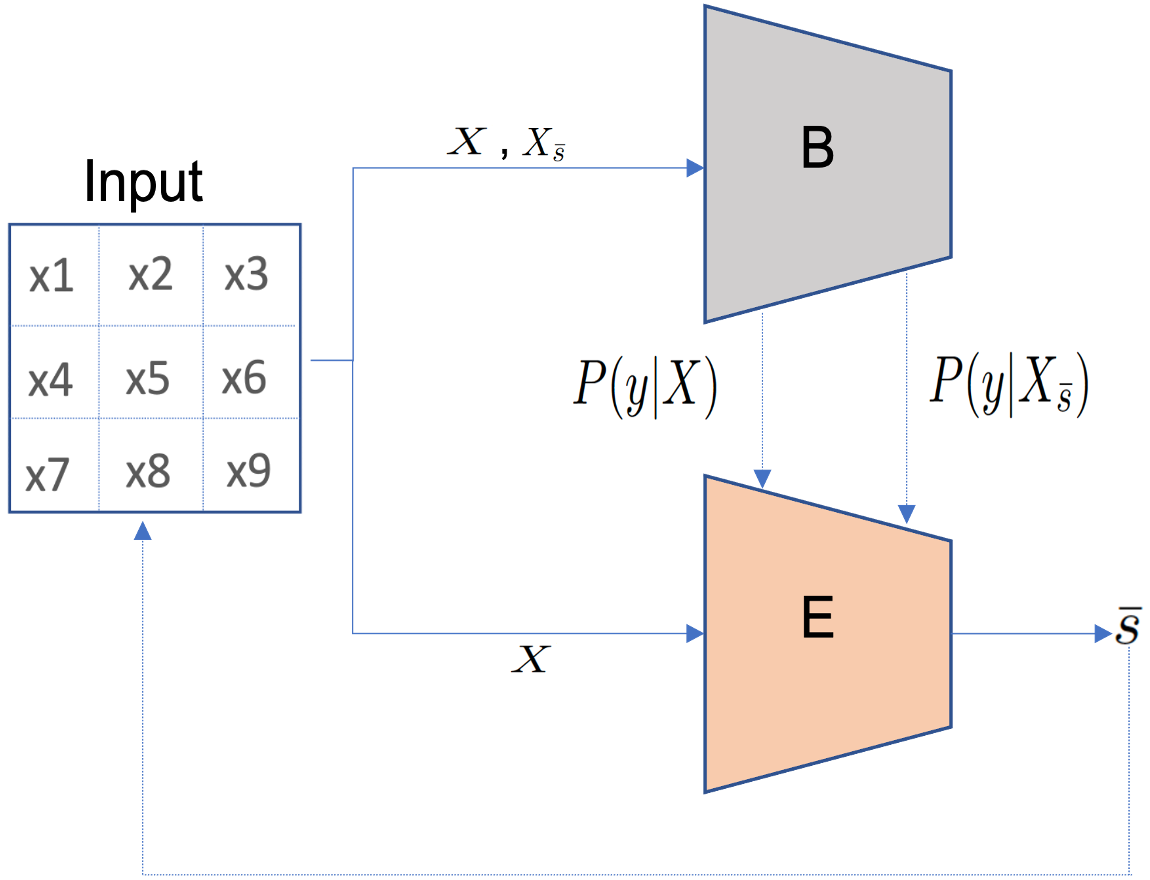}
    \caption{Casual feature selection based post-hoc explainer~(E)~training based on output of a pre-trained black-box~(B)}
    \label{fig:img1}
\end{figure}

\begin{figure*}[htpb]
    \centering
    \includegraphics[width=0.8\textwidth]{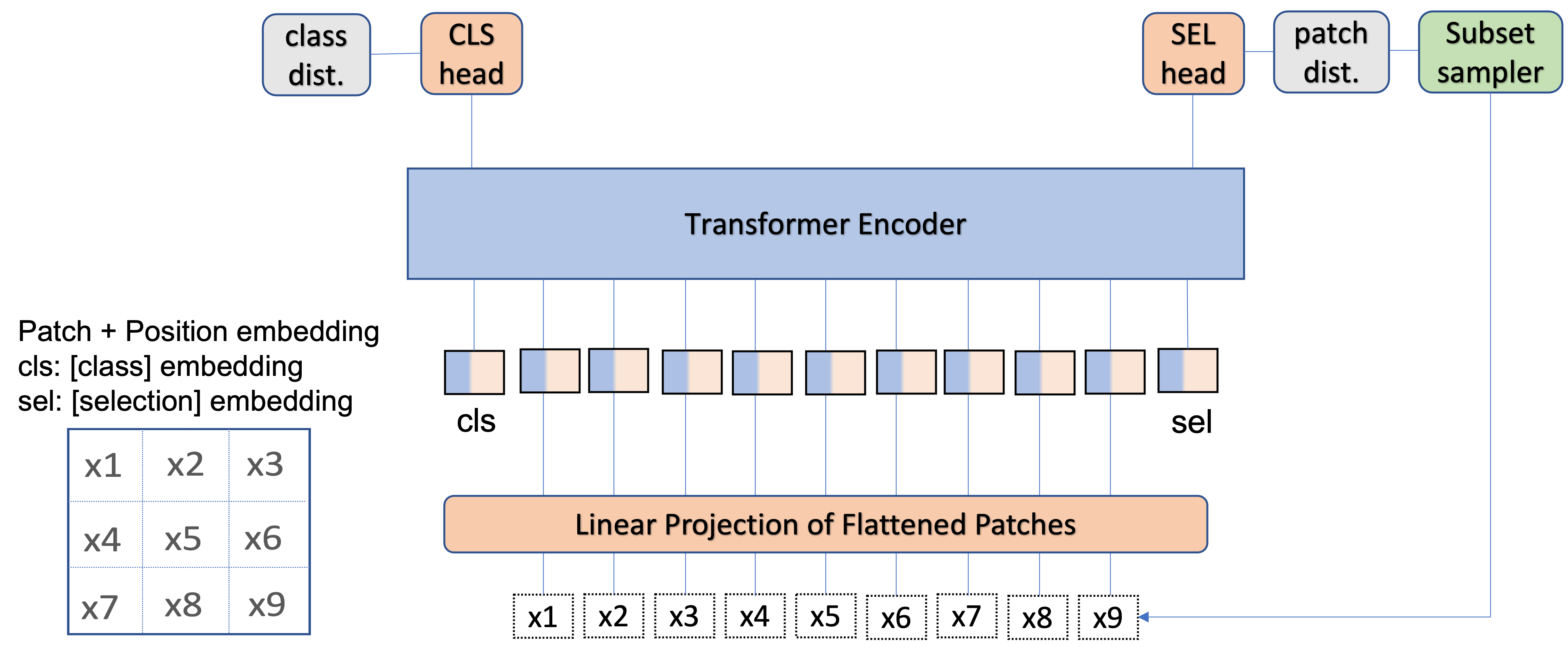}
    \caption{Our proposed expViT architecture with the continuous subset sampling based causal feature selection}
    \label{fig:img2}
\end{figure*}
Causality-based interpretation methods draw motivation from cognitive psychology of human reasoning~\cite{cognitive_psychology} and are accepted as the unifying approach for interpretability~\cite{covert2021explaining}.%
In a recent work by Panda~\etal~\cite{panda2021instance}, instance-wise causal feature selection is proposed. Their approach explains a given black-box model~($B$)~through a selector network~($E$)~trained to produce a categorical distribution from which a fixed set of features~($s$)~is sampled. To back-propagate through this network, the sampling operation should be differentiable. This is achieved by the continuous subset sampler built using the Gumbel-Softmax trick.~$E$~is trained with the objective function that maximizes the likelihood of the patches selected by it. For a given input image~($X$)~of class~$y$, a set of patches~($s$)~selected by~$E$~and the black box model~$B$, the loss function to train~$E$~as introduced in~\cite{panda2021instance} is as follows:
\begin{equation}
L_{\mathrm{sel}} = \sum\nolimits_{y=1}^{C} P(y|X) log(P(y|X_{\bar{s}}))
\label{eq:selection_loss}
\end{equation}
where, $X_{\bar{s}}$ is the input image with the set of patches selected by~$E$~zeroed out. $P(y|X)$ and $P(y|X_{\bar{s}})$ are distribution across $C$ classes, obtained from the black-box model for inputs $X$ and $X_{\bar{s}}$ respectively.

\section{Inherently explainable transformer}
\label{sec:Inherently explainable transformer}
We hypothesize that instead of building a separate explainer model, we can build an inherently explainable model. For this, we bring the causal feature selection approach developed in~\cite{panda2021instance} into the training of a transformer. The approach in~\cite{panda2021instance} assumes that there is a causal relationship between the input space and output space. Therefore, by simulating the breakage of this causality, we can identify the top-$k$ input regions contributing to the black-box model's decision. However, this assumption does not take into account relationships within the input space. Transformers, which are built from a series of self-attention layers leverage such relationships, hence should be able to better identify causally important regions. This motivated us to choose vision transformer (ViT)~\cite{dosovitskiy2020image} as our base model. 
\subsection{Explainable ViT}
ViT consumes an image as a sequence of flattened patches through a linear projection layer. The linearly projected patches are concatenated with their respective positional embeddings and passed through the transformer encoder, along with a learnable class embedding~($cls$). Into the existing ViT architecture, we include an additional learnable embedding token $sel$ which is passed through the transformer encoder. The output embedding corresponding to $sel$ is used by the selection head to produce probability distribution across the total number of patches which is then used to sample the $k$ patches most important for the model's decision. This inherently explainable ViT~($expViT$)~is trained by a loss function which is a weighted~($\lambda$)~sum of the standard cross-entropy~($\mathrm{CE}$)~loss for the classification and the selection loss as defined in \cref{eq:selection_loss}:
\begin{equation}
Loss = \mathrm{\lambda}*\mathrm{CE} + (1-\mathrm{\lambda})*L_\mathrm{sel}
\label{eq:expViT_loss}
\end{equation}

\subsection{Evaluation metrics}
Two causality based interpretation metrics used in~\cite{panda2021instance} are: post-hoc accuracy ($\mathrm{PA}$), defined as:
\begin{equation}
 \mathrm{PA} =\frac{1}{|X_{\mathrm{T}}|}\sum_{x \in X_{\mathrm{T}}} \mathbb{1}\big(\operatorname*{argmax}_y (P(y|x) = \operatorname*{argmax}_y (P(y|x_{s})\big)
\label{eq:ph_acc}
\end{equation}
 and, Average Causal Effect~($\mathrm{ACE}$)~, defined as:
 \begin{equation}
 \mathrm{ACE} = \frac{1}{|X_{T}|}\sum_{x \in X_{T}} (P(y|x_{s})- P(y|x_{rand}))
\label{eq:ace}
\end{equation}
Here, $X_{T}$ refers to the test set, and $x_{s}$ refers to image instances where the top $k$ patches from image $x$ are sampled from the learned categorical distribution, whereas for $x_{rand}$, $k$ patches are sampled from a uniform random distribution.  \cref{eq:ph_acc} computes how often the prediction of the model matches when it is passed the full input image as compared to the input image with only the top selected $k$ patches retained while zeroing out the rest. \cref{eq:ace} evaluates the causal strength of the image regions selected by the explainer compared to those selected at random.

\subsection{Experiments}
Similar to~\cite{panda2021instance}, for binary classification, we use a subset of the classes $(3,8)$,~(t-shirt and shoe), and~(bird and truck)~for MNIST, FMNIST, and CIFAR datasets respectively. A validation set~(20\%)~is randomly split
from the original training set. For post-hoc experiments, we use ViT for both black-box as well as selector models. However, the final layer of black-box ViT has only two neurons whereas the final layer of the selector ViT has neurons equal to the total number of $4*4$ sized patches in the input image. We first tune for hyper-parameters: $depth$ and $dim$ to train a black-box ViT. For which, we found the best $dim$ to be $512$ and the best $depth$ to be $6$, $4$, and $8$ achieving test-set accuracy of $0.993$, $0.999$, and $0.895$ for MNIST, FMNIST, and CIFAR dataset, respectively. Same dataset-specific hyper-parameters are then used to train both post-hoc selector ViT as well as expViT. For expViT, tuning of hyper-parameter $\lambda$ is also carried out separately for each fraction of input patches ($\mathrm{frac}$). All models are trained for $10$ epochs with a learning rate of $0.0001$ and $Adam$ optimizer. A comparison of the performance of the post-hoc selector ViT with our proposed expViT for different values of $\mathrm{frac}$ is presented in the results. For expViT, we also include the accuracy of the model when full input is passed ($\mathrm{ACC}$). Note that for expViT, evaluation metrics similar to post-hoc are computed by using $s$ patches selected by the $sel$ head of expViT.

\section{Results}

\begin{table}[htpb]
\centering
\begin{tabular}{c|cc|c|ccc}
\hline
Method&
\multicolumn{2}{|c|}{post-hoc}&
\multicolumn{4}{|c}{expViT} \\       
\hline
$\mathrm{frac}$    & $\mathrm{PA}$ & $\mathrm{ACE}$   & $\lambda$& $\mathrm{PA}$ & $\mathrm{ACE}$   & $\mathrm{ACC}$    \\
\hline
0.05      & 0.744   & 0.215 &  0.9 & \textbf{0.936}  & \textbf{0.422} & 0.968  \\
0.1       & 0.891   & 0.332 & 0.7 & \textbf{0.953}    &  \textbf{0.430} & 0.983  \\
0.25      & 0.952   & 0.302 & 0.6 & \textbf{0.969}   &  \textbf{0.415} & 0.982  \\
0.5       & \textbf{0.969}  & 0.196 & 0.7 & 0.968   &  \textbf{0.318} & 0.986  \\
\hline
\end{tabular}
\captionsetup{justification=centering}
  \caption{Post-hoc explainer ViT vs. expViT trained on MNIST subset. Black-box model had ACC of $0.993$}
  \label{Table:MNIST_results}
\end{table}

\begin{table}[htpb]
\centering
\begin{tabular}{c|cc|c|ccc}
\hline
Method&
\multicolumn{2}{|c|}{post-hoc}&
\multicolumn{4}{|c}{expViT} \\       
\hline
$\mathrm{frac}$ & $\mathrm{PA}$ & $\mathrm{ACE}$   & $\lambda$& $\mathrm{PA}$ & $\mathrm{ACE}$   & $\mathrm{ACC}$    \\
\hline
0.05      & 0.871   & 0.308 & 0.7& \textbf{0.970}    & \textbf{0.462} & 0.997  \\
0.1       & \textbf{0.992}   & 0.389 & 0.9 & 0.991   & \textbf{0.472} & 0.997  \\
0.25      & \textbf{0.995}   & 0.195 & 0.5 & 0.992   & \textbf{0.449} & 0.994  \\
0.5       & 0.986   & 0.033 & 0.6 & \textbf{0.987}   & \textbf{0.316} & 0.992  \\

\hline
\end{tabular}
\captionsetup{justification=centering}
  \caption{Post-hoc explainer ViT vs. expViT trained on FMNIST subset. Black-box model had ACC of $0.999$}
  \label{Table:FMNIST_results}
\end{table}

\begin{table}[htpb]
\centering
\begin{tabular}{c|cc|c|ccc}
\hline
Method&
\multicolumn{2}{|c|}{post-hoc}&
\multicolumn{4}{|c}{expViT} \\       
\hline
$\mathrm{frac}$ & $\mathrm{PA}$ & $\mathrm{ACE}$   & $\lambda$& $\mathrm{PA}$ & $\mathrm{ACE}$   & $\mathrm{ACC}$    \\
\hline
0.05      & \textbf{0.702}     & 0.095  &0.9   & 0.700   & \textbf{0.171} & 0.825  \\
0.1       & 0.779    & 0.122 & 0.9& \textbf{0.808}   & \textbf{0.280}  & 0.832  \\
0.25      & 0.774   & 0.134 & 0.9 & \textbf{0.820}    & \textbf{0.275} & 0.830   \\
0.5       & 0.778    & 0.130 & 0.9 & \textbf{0.832}   & \textbf{0.269} & 0.847  \\

\hline
\end{tabular}
    \captionsetup{justification=centering}
  \caption{Post-hoc explainer ViT vs. expViT trained on CIFAR subset. Black-box model had ACC of $0.895$}
  \label{Table:CIFAR_results}
\end{table}

\begin{figure}[htbp]
    \centering
    \includegraphics[width=\linewidth]{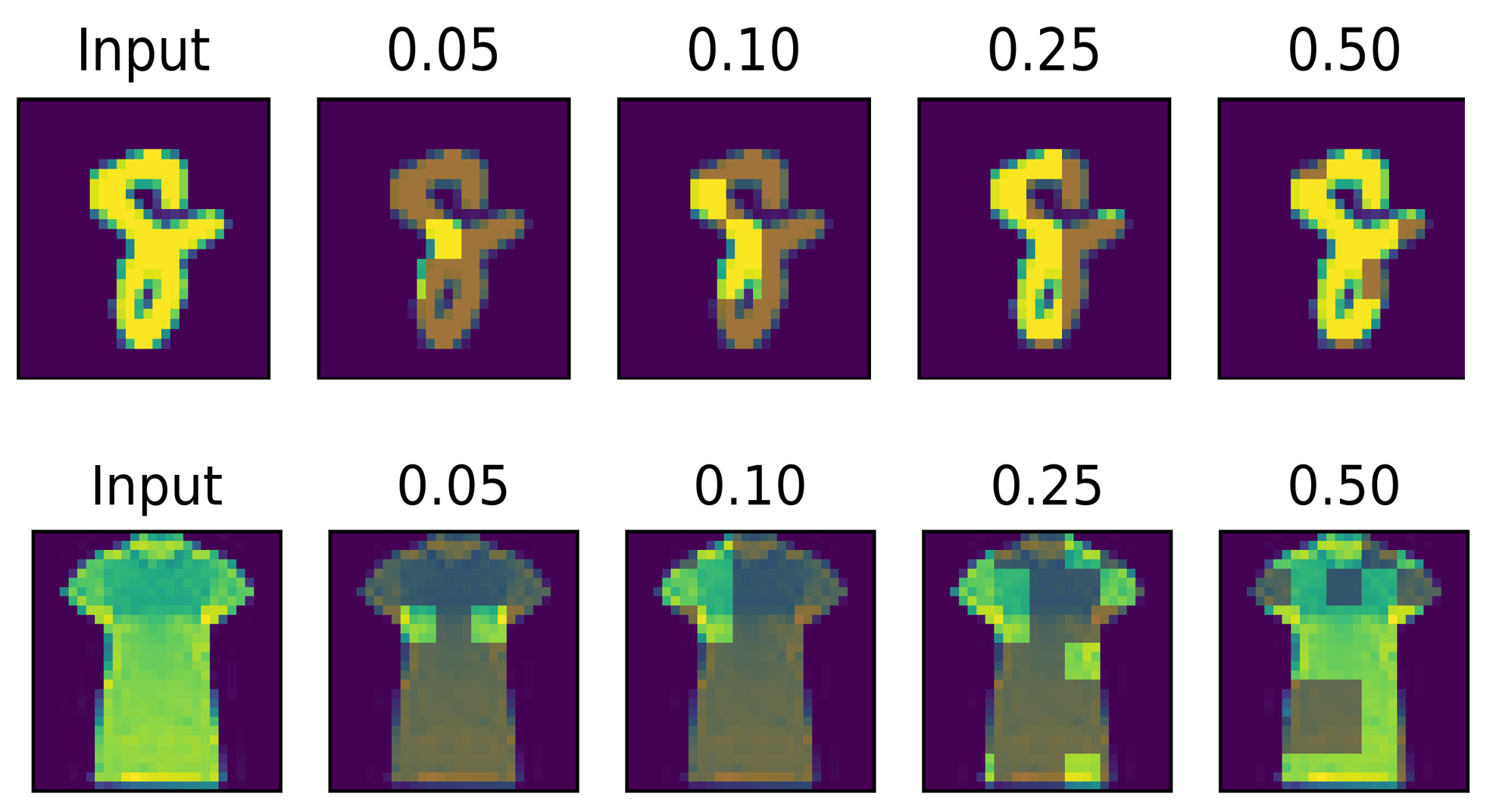}
    \caption{Qualitative results of expViT trained on MNIST subset (digits: 3 and 8) and FMNIST subset (t-shirt and shoe) for different fractions of top causal patches selected}
    \label{fig:img3}
\end{figure}
\section{Conclusion}
As evident from the results, our inherently explainable model mostly performs better than the post-hoc explainer on two explainability metrics. However, there is a trade-off between explainability and the true accuracy of the model with full input ($\mathrm{ACC}$). This trade-off for the proposed expViT can be tuned based on two parameters: $\lambda$ in \cref{eq:expViT_loss} and $\mathrm{frac}$ of the total patches desired to be selected.
%%%%%%%%% REFERENCES
{\small
\bibliographystyle{ieee_fullname}
\bibliography{CAT-XPLAIN}
}
\let\thefootnote\relax\footnotetext{This paper was accepted for spotlight presentation at the Explainable Artificial Intelligence for Computer Vision Workshop at CVPR 2022.}
\end{document}